\documentclass[letterpaper, 10 pt, conference]{ieeeconf}
\IEEEoverridecommandlockouts
\usepackage{graphicx} 
\usepackage{soul}
\usepackage{color, xcolor}
\usepackage{amsmath}
\usepackage{amsfonts}
\usepackage{mathrsfs}
\usepackage{graphicx}
\usepackage{booktabs}
\usepackage{float}
\usepackage{booktabs}
\usepackage{multirow}
\usepackage{threeparttable}
\usepackage{array}
\usepackage{subfig}
\usepackage{float}
\usepackage{url}
\usepackage{diagbox}
\title{\LARGE 
VISO: Robust Underwater Visual-Inertial-Sonar SLAM with Photometric Rendering for Dense 3D Reconstruction}

\author{Shu Pan$^{1*}$, Simon Archieri$^{1}$, Ahmet Cinar$^{2}$, Jonatan Scharff Willners$^{2}$, \\ Ignacio Carlucho$^{1}$ and Yvan Petillot$^{1}$
\thanks{$^{1}$ School of Engineering and Physical Sciences, Heriot-Watt University, Edinburgh, UK}
\thanks{$^{2}$ Frontier Robotics, The National Robotarium, Edinburgh, UK}
}

\begin{document}




\maketitle

\begin{abstract}
Visual challenges in underwater environments significantly hinder the accuracy of vision-based localisation and the high-fidelity dense reconstruction. In this paper, we propose VISO, a robust underwater SLAM system that fuses a stereo camera, an inertial measurement unit (IMU), and a 3D sonar to achieve accurate 6-DoF localisation and enable efficient dense 3D reconstruction with high photometric fidelity. We introduce a coarse-to-fine online calibration approach for extrinsic parameters estimation between the 3D sonar and the camera. Additionally, a photometric rendering strategy is proposed for the 3D sonar point cloud to enrich the sonar map with visual information. Extensive experiments in a laboratory tank and an open lake demonstrate that VISO surpasses current state-of-the-art underwater and visual-based SLAM algorithms in terms of localisation robustness and accuracy, while also exhibiting real-time dense 3D reconstruction performance comparable to the offline dense mapping method.
\end{abstract}


\section{Introduction}\label{sec:intro}
Underwater simultaneous localisation and mapping (SLAM) is essential for a wide range of tasks, including environmental monitoring, offshore infrastructure inspection, marine archaeology, and autonomous manipulation. SLAM provides underwater vehicles with both accurate pose estimation and reliable environmental perception. 
%
%
However, the underwater environment presents unique characteristics that make accurate localisation and high-fidelity 3D reconstruction challenging. First, sensors such as GPS and Lidars, which are widely used on the ground domain, are unavailable underwater.
In addition, visual sensing is severely degraded by light attenuation, scattering, and colour distortion, particularly in turbid waters \cite{rahman2019svin2}. Although multi-beam sonars such as Forward-Looking Sonar (FLS) are unaffected by turbidity, they capture only 2D images, leading to 3D positional ambiguity and pose significant challenges for 3D mapping \cite{hurtos2015fourier,hansen2023using}.

To address these challenges, multi-modal sensor fusion strategies 
have been extensively explored in existing underwater SLAM methods. 
Cameras, which serve as essential sensors in underwater inspection by providing rich visual information content, have been integrated with other sensing modalities. In particular, visual–inertial systems have been fused with Doppler Velocity Logs (DVLs) \cite{xu2021underwater, xu2025aqua}, profiling sonars \cite{rahman2019svin2, rahman2022svin2}, and imaging sonars \cite{collado2025opti, pan2025russo}, \cite{singh2024opti} to achieve robust underwater localisation and mapping. These solutions can indeed improve the robustness and accuracy of localisation, however, mapping still heavily relies on camera visibility, making 3D scene reconstruction a challenging problem in turbid environments. In contrast, FLS sonars are less susceptible to visually challenging conditions and have been fused with Inertial Measurement Unit (IMU) and DVLs to achieve accurate perception in murky underwater environments \cite{li2018pose}, \cite{wang2022virtual}, \cite{xu2024diso}, \cite{archieri20253dssdf}. However, images captured by FLS suffer from degradation in elevation angle, causing multi-modal SLAM systems to still struggle with full 6-DoF pose estimation and 3D reconstruction.

\begin{figure}
  \centering
  \includegraphics[width=0.48\textwidth]{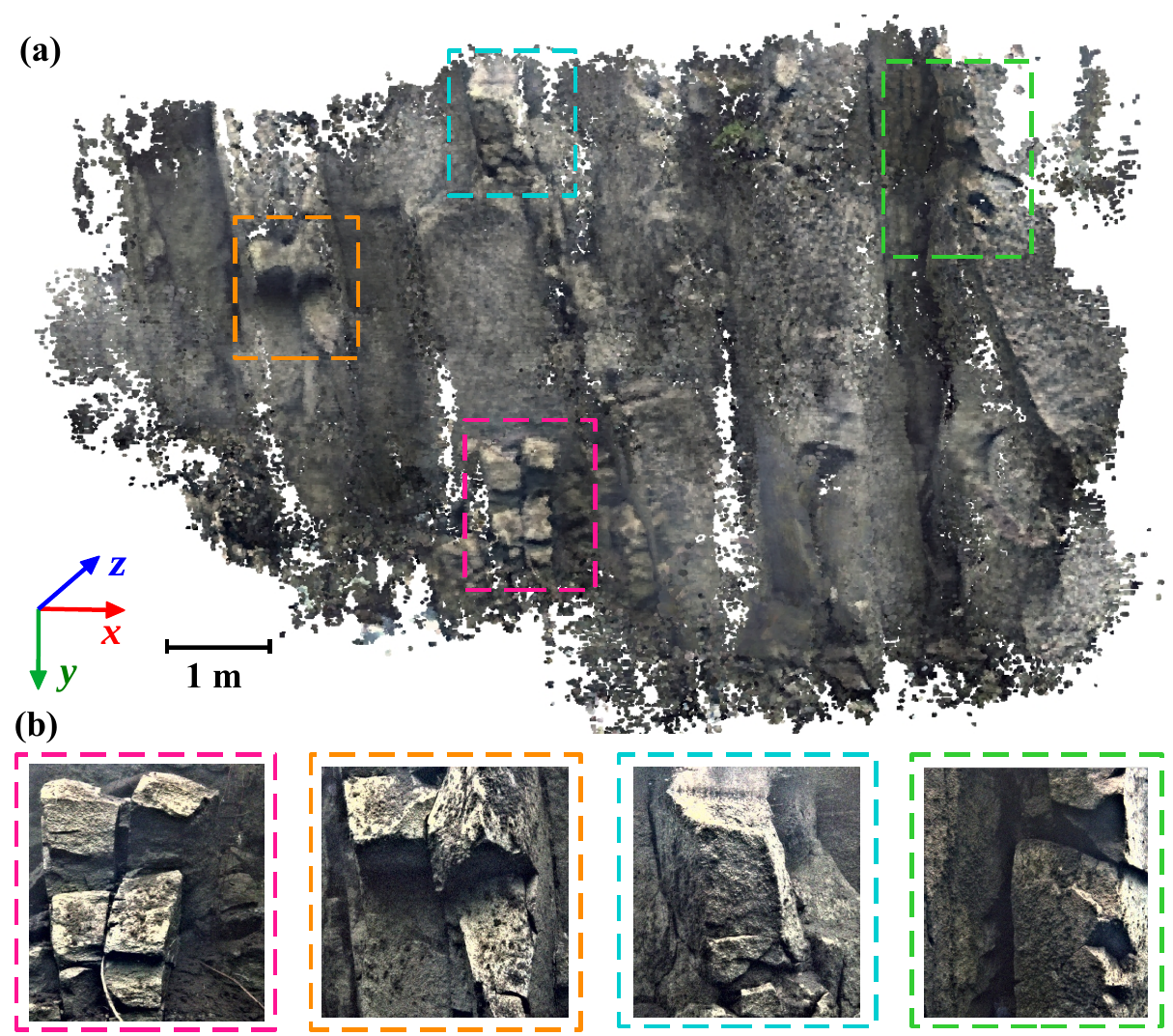}
  \caption{(a) Dense mapping result in the lake. (b) The corresponding camera view of colour-dotted boxes in the areas of interest on the dense sonar map.}
  \label{fig:LakeMapping}
\end{figure}
\begin{figure*}[t!]
    \centering
    \includegraphics[width=0.95\linewidth]{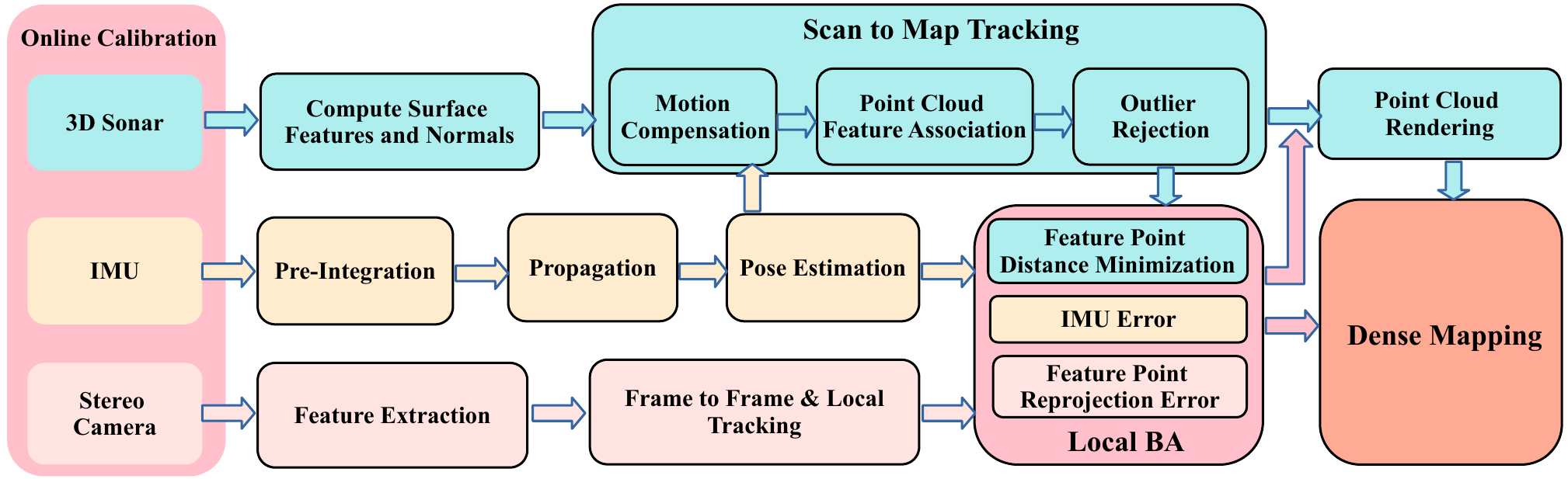}
    \caption{Overview of VISO, where a 3D sonar is fused with an IMU and a stereo camera to enable accurate localisation and real-time dense mapping. }
    \label{fig:system overview}
\end{figure*}
In this paper, we present a robust and accurate underwater Visual-Inertial-Sonar SLAM system (\textbf{VISO}) that incorporates an underwater 3D sonar,  with a stereo camera, and an IMU to achieve full 6-DoF localisation and real-time 3D dense mapping with high photometric fidelity in underwater environments. 
%
Specifically, we fuse the sparse point clouds provided by the 3D sonar with camera and IMU measurements in a tightly coupled framework to jointly optimise 6-DoF pose estimation. Moreover, the 3D sonar data is effectively combined with the rich visual information from the camera to enable real-time dense 3D reconstruction with photometric rendering. The main contributions of this work are summarised as follows:
\begin{enumerate}
    \item We propose an online calibration approach for estimating the extrinsic parameters between the 3D sonar and camera without requiring any prior assumptions;
    \item We present an accurate 3D sonar point cloud association method with outlier rejection to address the challenges of sparsity and noise;
    \item We introduce a novel dense mapping solution for real-time underwater 3D reconstruction with photometric rendering of 3D sonar point clouds;
    \item We conduct extensive experiments in both a laboratory tank and a lake, demonstrating the superiority of VISO in underwater localisation and its effectiveness in dense reconstruction with high photometric fidelity.
\end{enumerate}

The remainder of this paper is organised as follows. Related work is discussed in Section \ref{sec:related}. Section \ref{sec:approach} provides an overview of our proposed visual-inertial-sonar SLAM framework and the details of the algorithm. Experimental results are presented in Section \ref{sec:experiment}. Finally, conclusions are drawn in Section \ref{sec:conclusion}.
\section{RELATED WORK}\label{sec:related}
\subsection{3D Sonar-Camera Extrinsic Calibration}
3D sonar–camera calibration has been rarely explored, \cite{marburg2015extrinsic} is the only work that proposed an extrinsic calibration method between a 3D imaging sonar and an RGB camera using a specific object setup in a laboratory pool, which is complex and time-consuming. 3D sonar shares similar characteristics with LiDAR, and several works on LiDAR–camera calibration have been proposed in \cite{park2020spatiotemporal}, \cite{yuan2021pixel}. However, 3D sonar point clouds are considerably more sparse and noisy than LiDAR point clouds, as shown in Fig.~\ref{fig:SonarCloudShow}, and are closer to the 4D radar point clouds. While several works have proposed different 4D radar-camera calibration approaches to obtain the extrinsic parameters  \cite{zhuang20254drc}, \cite{cao20254d}, the significant differences in the measurement range and field of view (FOV) between 3D sonar and 4D radar hinder the direct application of these methods to 3D sonar–camera calibration.

\subsection{Sonar-based Underwater SLAM}
Sonars have been widely used for underwater SLAM. Approaches that rely solely on imaging sonar for underwater SLAM have been proposed in \cite{hurtos2015fourier} and \cite{hansen2023using} to achieve localisation and mosaic generation. To achieve a more robust perception, 
\cite{li2018pose},
\cite{wang2022virtual}, \cite{xu2024diso}, \cite{archieri20253dssdf}
fused imaging sonar, IMU, and DVL to enable robust localisation and mapping. However, imaging sonar data degrade in elevation angle, posing significant challenges for full 6-DoF pose estimation and 3D mapping. To address this issue,  \cite{mcconnell2022overhead}, \cite{mcconnell2024large} fused two orthogonal imaging sonar datasets to solve elevation ambiguity and achieve large-scale underwater localisation and 3D reconstruction. Nevertheless, these methods face difficulties in orthogonal sonar data association and still struggle to achieve accurate 6-DoF localisation. 3D sonars are a promising sensor for addressing these challenges. Prior works \cite{ferreira20223dupic}, \cite{ferreira2025real}, and \cite{hansen2005mosaicing} have investigated underwater SLAM using only 3D sonar. However, relying solely on 3D sonar remains fragile due to the inherent sparsity and noise of the data.

\subsection{Visual-based Underwater SLAM}
Cameras play a crucial role in underwater inspection and have been fused with other sensing modalities in visual-based underwater SLAM frameworks. A robust underwater SLAM that fuses a stereo camera, an IMU, and a profiling sonar has been proposed to address visual challenges in underwater localisation \cite{rahman2019svin2}, \cite{rahman2022svin2}. Recently, imaging sonar has been used to integrate with a visual–inertial system to mitigate visual degradation \cite{pan2025russo}. While \cite{collado2025opti} proposed fusing segmented camera images with sonar data to enable localisation and 3D reconstruction in highly turbid underwater environments. In addition, cameras have been combined with other sensing modalities such as DVL, barometers, gyroscopes, and pressure sensors to achieve more robust underwater localisation and mapping \cite{xu2021underwater}, \cite{xu2025aqua}, \cite{song2024turtlmap}, \cite{huang2023tightly}, \cite{vargas2021robust}. However, the unique underwater visual challenges hinder the dense 3D reconstruction capability of current visual-based underwater SLAM systems.

\begin{figure}[h]
  \centering
  \includegraphics[width=0.48\textwidth]{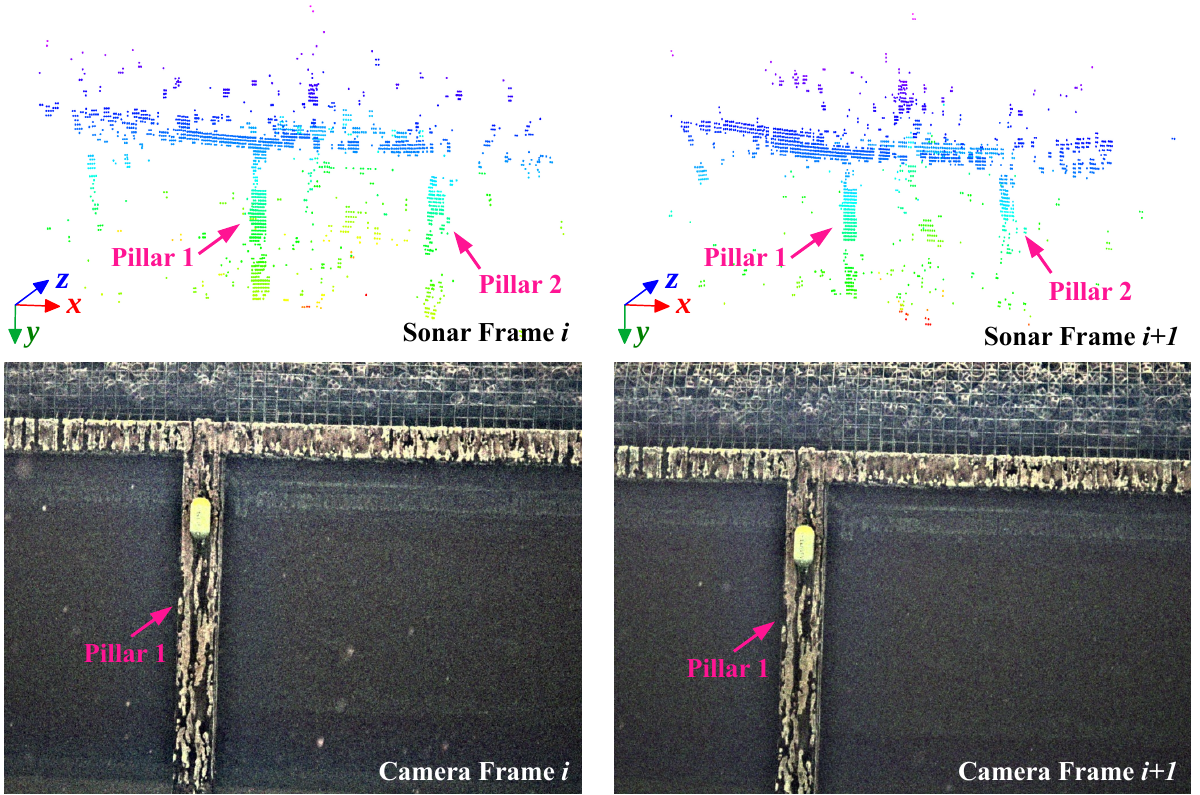}
  \caption{The visualisation of sparse and noisy sonar point clouds with corresponding camera images in different viewpoints. The point clouds vary in different perspectives.}
  \label{fig:SonarCloudShow}
\end{figure}

\section{Method}\label{sec:approach}
In this work, we propose a robust underwater SLAM system, VISO, which fuses a stereo camera, an IMU, and a 3D sonar to achieve accurate full 6-DoF localisation and real-time dense mapping with photometric rendering in underwater environments. An overview of the SLAM system is shown in Fig.~\ref{fig:system overview}.

\subsection{Notations}
We adopt the following coordinate frame definitions: $(\cdot)^w$, $(\cdot)^i$, and $(\cdot)^k$ denote the 3D points in the world frame, current frame, and key frame, respectively. $\mathbf{T}_{WC}$, $\mathbf{T}_{WSo}$, $\mathbf{T}_{WI} = [\mathbf{q}_{wb} \mid \mathbf{p}_{wb}]$ $\in$ $\mathbf{SE}(3)$ represent the camera, 3D sonar, IMU (body) pose in world frame,  respectively. The robot state is defined as $\mathbf{x}_R= [\mathbf{q}^{T}_{wb},\mathbf{p}^{T}_{wb},\mathbf{v}^{T}_{wb},\mathbf{b}^{T}_\omega,\mathbf{b}^{T}_a]^T \in SO(3) \times \mathbb{R}^3 \times \mathbb{R}^9$, where $\mathbf{b}_\omega,\mathbf{b}_a$ represent the gyroscopes and accelerometers bias, respectively. Both $\mathbf{q}_{wb}$ and $\mathbf{R}_{wb}$ are adopted to represent rotation. $\mathbf{g}^{w} = \begin{bmatrix}
    0, 0, g
\end{bmatrix}^{T}$ is the gravity vector in the world frame.

\subsection{Online Extrinsic Calibration}
The transformation $\mathbf{T}_{IC}$ from the camera frame to the IMU frame, and the transformation $\mathbf{T}_{CSo}$ from the 3D sonar frame to the camera frame are crucial for tightly-coupled pose estimation. First, $\mathbf{T}_{IC}$ is calibrated using an online calibration method \cite{qin2017vins}. Then $\mathbf{T}_{CSo}$ is calibrated through the following two stages:
\subsubsection{Coarse Calibration}
We first estimate a coarse transformation $\hat{\mathbf{T}}_{CSo}$ using the initial $n$ camera and 3D sonar poses, where the camera poses are obtained from the visual–inertial system, and the 3D sonar poses are estimated via a coarse 3D sonar odometry, as follows.

Given the current and keyframe 3D sonar point cloud $\mathbf{P}_{So_i}$ and $\mathbf{P}_{So_k}$, the pose transformation $\mathbf{T}_{So_kSo_i}$ from current frame to keyframe can be estimated by:
\begin{equation}
\mathbf{P}_{So_k} = \mathbf{T}_{So_k So_i}\mathbf{P}_{So_i}
\label{eq:sonar}.
\end{equation}
Then, the continuous 3D sonar pose can be updated based on the key sonar frame pose $\mathbf{T}_{So_k}$ according to:
\begin{equation}
\mathbf{T}_{So_i} = \mathbf{T}_{So_k}\mathbf{T}_{So_k So_i}
\label{eq:sonar}.
\end{equation}
After that, a set of camera poses $\mathbf{T}_{WC_i}$ and 3D sonar poses $\mathbf{T}_{So_i}$, $i \in [1, \mathcal{N}]$ are obtained and used to formulate the relative pose constraint, which is defined as:
\begin{equation}
\mathbf{T}_{C_{i-1}C_i} \mathbf{\hat{T}}_{CSo}= \mathbf{\hat{T}}_{CSo} \mathbf{T}_{So_{i-1}So_i}.
\label{eq:sonar}
\end{equation}
Finally, the coarse extrinsic $\mathbf{\hat{T}}_{CSo}$ is optimized by solving the following problem using nonlinear optimization:
\begin{equation}
\underset{\mathbf{\hat{T}}_{CSo}}{\operatorname{argmin}} = \sum_{i=1}^{\mathcal{N}} || \mathbf{T}_{C_{i-1}C_i} \mathbf{\hat{T}}_{CSo} - \mathbf{\hat{T}}_{CSo} \mathbf{T}_{So_{i-1}So_i} ||^2.
\label{eq:sonar}
\end{equation}
\subsubsection{Refined Calibration}
The coarse extrinsic estimation can not be sufficiently accurate due to errors in the coarse 3D sonar odometry, which are caused by the sparse and noisy nature of 3D sonar measurements, as well as the significant variation in point clouds across different viewpoints, as illustrated in Fig.~\ref{fig:SonarCloudShow}. Therefore, a refined calibration is required.
In this stage, we focus on registering camera landmarks with the sonar point cloud. Given the current sonar point cloud $\mathbf{P}_{So_i}$, it can be transformed to the world frame as $\mathbf{P}^w_{So_i}$ using the coarse calibration result:
\begin{equation}
\mathbf{P}^w_{So_i} = \mathbf{T}_{WC_i}\mathbf{\hat{T}}_{CSo}\mathbf{P}_{So_i}
\label{eq:sonar}.
\end{equation}
Next, we search for points in $\mathbf{P}^w_{So_i}$ that are close to the camera landmarks $\mathbf{P}^w_C$, retaining only those within a specified radius $\mu$. The selected points form the subset $\mathbf{\hat{P}}^w_C$, and the two sets of points are then aligned as:
\begin{equation}
\mathbf{P}^w_C = \mathbf{T}_{P2P}\mathbf{\hat{P}}^w_C
\label{eq:sonar}.
\end{equation}
Finally, the refined extrinsic transformation between the 3D sonar and the camera is obtained as:
\begin{equation}
\mathbf{T}_{CSo} = \mathbf{T}_{WC_i}^{-1}\mathbf{T}_{P2P}\mathbf{T}_{WC_i}\mathbf{\hat{T}}_{CSo}
\label{eq:sonar}.
\end{equation}

\begin{figure}
  \centering
  \includegraphics[width=0.48\textwidth]{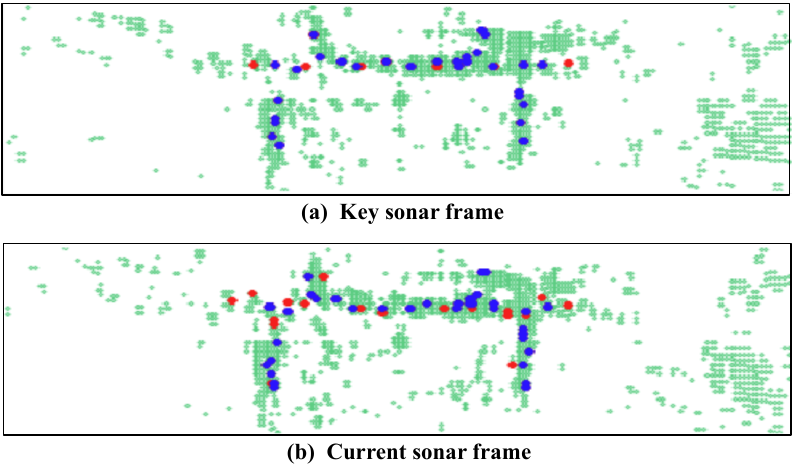}
  \caption{The key sonar frame (a) and the current sonar frame (b) data association results visualised in the XY plane. {\color{green}{Green}} points are the backprojected 3D sonar point cloud, {\color{red}{red}} points are the outliers after initial matching, {\color{blue}{blue}} points are the matched features after outlier rejection.}
  \label{fig:sonarMatching}
\end{figure}

\subsection{3D Sonar Data Association and Residual}
\subsubsection{\textbf{Compute Surface Features and Normals}}
Inspired by \cite{adolfsson2022lidar} and \cite{wu2024efear}, we first partition the 3D sonar point cloud into $\mathcal{V}$ voxels. For each voxel, we search for its neighbouring points, which include both points inside the voxel and adjacent points from neighbouring voxels. If the number of neighbouring points exceeds a threshold $\gamma$, the voxel is selected as a surface feature, and represented using the mean of its points as $\mathbf{P}_m$. Finally, the normal vector $\mathbf{u}_m$ of the voxel is calculated as the description of the voxel using principal component analysis (PCA). 

\subsubsection{\textbf{Scan to Map Tracking}}
Once the surface features and normals are obtained, scan-to-map tracking is performed to associate the features in the current frame and recent keyframes. First, a motion prior $\mathbf{\hat{T}}_{So_kSo_i}$, provided by IMU propagation, is used to transform the current 3D sonar frame $\mathbf{P}_{So_i}$ into the previous $\mathcal{K}$ keyframes as:
\begin{equation}
\mathbf{\hat{P}}_{So_k} = \mathbf{\hat{T}}_{So_kSo_i}\mathbf{P}_{So_i} 
\label{eq:sonar}.
\end{equation}
Next, each voxel in the transformed frame $\mathbf{\hat{P}}_{So_k}$ searches for its corresponding voxel in the keyframe within a radius $\gamma$. A voxel correspondence is established if the following conditions are satisfied:
\begin{equation}
    \begin{cases}
        ||\mathbf{P}^i_m - \mathbf{P}^k_n|| < \gamma\\
        \mathbf{u}^i_m\cdot\mathbf{u}^k_n > l
    \end{cases},
\end{equation}
where $l$ represent the similarity of two surface normals, and $k \in [1, \mathcal{K}]$, $m, n \in [1, \mathcal{V}]$.

\subsubsection{\textbf{Outlier Rejection}}
As a result, we obtain a set of matched 3D sonar feature points $(\mathbf{\hat{P}}^i, \mathbf{\hat{P}}^k)$. However, these associations may not be sufficiently accurate, especially in environments with highly similar structures, where the normal vectors can be nearly identical. As shown in Fig.~\ref{fig:sonarMatching}, some outliers (red points) remain after data association. Therefore, we perform outlier rejection using 2D–2D RANSAC on the back-projected points in the current and key sonar frames to refine the associations $(\mathbf{P}^i, \mathbf{P}^k)$, as indicated by the blue points.

\subsubsection{\textbf{Feature Point Distance Minimization Residual}}
The accurate associations are used as constraints to optimise the current robot pose $\mathbf{T}_{WI_i}$. Given the keyframe pose $\mathbf{T}_{WI_k}$, the 3D sonar feature–based distance error is defined as:
\begin{equation}
E_{so} = \mathbf{T}_{WI_k} \mathbf{T}_{ISo}\mathbf{P}^k - \mathbf{T}_{WI_i} \mathbf{T}_{ISo}\mathbf{P}^i
\label{eq:sonar},
\end{equation}
where $\mathbf{T}_{ISo}$ represents the transformation from sonar frame to IMU frame, which can be derived from the calibrated extrinsic.
\subsection{IMU Residual}
The raw gyroscope and accelerometer measurements from an IMU are given by:
\begin{equation}
\begin{aligned}
  \hat{\mathbf{a}}_{t} &=  \mathbf{a}_{t} + \mathbf{q}^t_{bw}\mathbf{g}^{w} + \mathbf{b}_{a_t} +  \mathbf{n}_{a}\\
  \hat{\boldsymbol{\omega}}_{t} &=  \boldsymbol{\omega}_{t} + \mathbf{b}_{\omega_t} +  \mathbf{n}_{\omega}\\
\end{aligned},
\end{equation}
where $\hat{\boldsymbol{\omega}}_{t}$, $\hat{\mathbf{a}}_{t}$ are the raw measurements in the body frame at time $t$, and they are affected by acceleration bias $\mathbf{b}_{a_t}$, gyroscope
bias $\mathbf{b}_{\omega_t}$, acceleration noise $\mathbf{n}_{a} \sim \mathcal{N}(0, \boldsymbol{\sigma}_a^2)$ and gyroscope noise $\mathbf{n}_{\omega}\sim \mathcal{N}(0, \boldsymbol{\sigma}_g^2)$. $\mathbf{q}^t_{bw}$ is the rotation from the world frame to the body frame.

Given the bias estimation, the inertial measurement over the interval $[t_k, t_{k+1}]$ can be preintegrated as follows:
\begin{equation}\label{pre-integration}
\begin{aligned}
\boldsymbol{\alpha}_{b_{k}b_{k+1}} = \int\int_{t\in[k, {k+1}]}\mathbf{q}^{b_k}_t(\mathbf{a}_{t}-\mathbf{b}_{a_t})d t^2 \\ 
\boldsymbol{\beta}_{b_{k}b_{k+1}} = \int_{t\in[k, {k+1}]}\mathbf{q}^{b_k}_t(\mathbf{a}_{t}-\mathbf{b}_{a_t})dt \\
\mathbf{q}_{b_{k}b_{k+1}} = \int_{t\in[k, {k+1}]}\mathbf{q}^{b_k}_t\otimes
\begin{bmatrix}
    0 \\
    \frac{1}{2}\boldsymbol{(\omega}_{t}- \mathbf{b}_{\omega_t}) \\
  \end{bmatrix}dt
\end{aligned} .
\end{equation}

For two consecutive frames $b_k$ and $b_{k+1}$, given the pose in frame $b_k$, then the position $\mathbf{p}_{wb_{k+1}}$, velocity $\mathbf{v}_{wb_{k+1}}$, and rotation $\mathbf{q}_{wb_{k+1}}$ in frame $b_{k+1}$ can be estimated by the IMU propagation using the IMU pre-integration terms by:
\begin{equation}
\begin{aligned}
  \mathbf{p}_{wb_{k+1}} &=    \mathbf{p}_{wb_k} + \mathbf{v}_{wb_k}\Delta t - \frac{1}{2}\mathbf{g}^w\Delta t^2 + \mathbf{q}_{wb_k}\boldsymbol{\alpha}_{b_{k}b_{k+1}}\\
  \mathbf{v}_{wb_{k+1}} &= \mathbf{v}_{wb_k} - \mathbf{g}^w \Delta t + \mathbf{q}_{wb_k}\boldsymbol{\beta}_{b_{k}b_{k+1}}\\
  \mathbf{q}_{wb_{k+1}} &=    \mathbf{q}_{wb_k}\mathbf{q}_{b_{k}b_{k+1}} \\
\end{aligned}.
\end{equation}

Then, the IMU pose error between consecutive frames $b_k$ and $b_{k+1}$ can be formulate by:
\begin{equation}
\begin{aligned}
 \mathcal{R}_{P} &= \resizebox{.87\linewidth}{!}{$
\mathbf{q}_{b_{k}w}(\mathbf{p}_{wb_{k+1}}-\mathbf{p}_{wb_k}-\mathbf{v}_{wb_k}\Delta t+\tfrac12\mathbf{g}^w\Delta t^2)-\boldsymbol{\alpha}_{b_{k}b_{k+1}}
$} \\
  \mathcal{R}_{V} &= \mathbf{q}_{b_{k}w}(\mathbf{v}_{wb_{k+1}} - \mathbf{v}_{wb_k} + \mathbf{g}^w \Delta t) -\boldsymbol{\beta}_{b_{k}b_{k+1}}\\
  \mathcal{R}_{Q} &= 2[(\mathbf{q}_{b_{k}b_{k+1}})^{-1} \otimes(\hat{\mathbf{q}}_{b_{k}w}\otimes\mathbf{q}_{wb_{k+1}})]_{xyz} \\
  \mathcal{R}_{b_\omega} &= \mathbf{b}_{\omega_{k+1}} - \mathbf{b}_{\omega_k} \\
  \mathcal{R}_{b_a} &= \mathbf{b}_{a_{k+1}} - \mathbf{b}_{a_k}
\end{aligned},
\end{equation}
where $[\cdot]_{xyz}$ means taking the vector part from a quaternion.

Finally, these error terms are used to formulate the IMU residual $E_{I}$, which can be expressed as:
\begin{equation}
E_{I} = 
\begin{bmatrix}
\mathcal{R}_{P},
 \mathcal{R}_{V},
 \mathcal{R}_{Q},
 \mathcal{R}_{b_w},
 \mathcal{R}_{b_a}
\end{bmatrix}^T.
\end{equation}

\subsection{Camera Residual}
The matched visual observations $\mathbf{z}^{s,j,k}$ in the $k^{th}$ key camera frame and their corresponding landmarks $\mathbf{P}_{C_{s,j}}$ in the current camera frame are then used to formulate the feature point reprojection error as:
\begin{equation}
E^{s,j,k}_C = \mathbf{z}^{s,j,k} - \pi_s(\mathbf{P}_{C_{s,j}}),
\end{equation}
where $s$ denotes the camera index of the stereo camera, $j$ is the index of the observation or its corresponding landmark, and $\pi_s(\cdot)$ is the camera projection model. In addition, the landmarks $\mathbf{P}_{C_{s,j}}$ can be represented using the current robot pose $\mathbf{T}_{WI}$, and the corresponding landmarks $\mathbf{P}^w_j$ in the world frame by:
\begin{equation}
\mathbf{P}_{C_{s,j}} = \mathbf{R}_{c_sb}\mathbf{R}_{wb_k}^{-1}(\mathbf{P}^w_j-\mathbf{p}_{wb_k})+\mathbf{p}_{c_sb},
\end{equation}
where $\mathbf{T}_{C_sI}=[\mathbf{R}_{c_sb}|\mathbf{p}_{c_sb}] \in \mathbf{SE}(3)$ donates the transformation from IMU to camera frame.

\subsection{Visual-Inertial-Sonar Joint Optimization}
Finally, the camera, IMU, and sonar residuals are jointly optimised in the local bundle adjustment (BA) module, which maintains a sliding window of up to $\mathcal{K}$ keyframes to enable real-time optimisation. The cost function is formulated using these three types of residuals:
\begin{equation}
\begin{split}
\mathbf{J}(\mathbf{X}) = &\sum_{k=1}^{\mathcal{K}}\sum_{j\in k}E_{So}^{j,k^T}\mathbf{P}_{So}^{k}E_{So}^{j,k^T} +\sum_{k=1}^{\mathcal{K}-1}E_{I}^{k^T}\mathbf{P}_{I}^{k}E_{I}^{k} \\ &+\sum_{n=1}^{N=2}\sum_{k=1}^{\mathcal{K}}\sum_{j\in\gamma(n,k)} E_{C}^{n,j,k^T}\mathbf{P}_{C}^{k}E_{C}^{n,j,k}\end{split},
\end{equation}
where $\mathbf{P}_{So}^{k}$, $\mathbf{P}_{I}^{k}$ and $\mathbf{P}_{C}^{k}$ are the information matrices of sonar observation, IMU, and camera landmarks, respectively.
\subsection{Sonar Cloud Rendering and Dense Mapping}
The 3D sonar point clouds $\mathbf{P}_{So_i}$ can be rendered using the optimised pose $\mathbf{T}_{WI_i}$ along with the corresponding camera images, and are then used to construct the dense map $\mathcal{M}$, which can be expressed as:
\begin{equation}
\begin{bmatrix}
    x^{i,j},
    y^{i,j},
    z^{i,j},
    g^{i,j}
  \end{bmatrix}^T = \mathcal{M}^{i,j} = \mathbf{T}_{WI_i} \mathbf{T}_{ISo}\mathbf{P}_{So_{i,j}},
\end{equation}
where $g^{i,j}$ denotes the colour of the $j^{th}$ point in the point cloud, obtained by projecting the point into the current camera image:
\begin{equation}
g^{i,j} = \mathcal{C}(\pi_s(\mathbf{T}_{WI_i} \mathbf{T}_{ISo}\mathbf{P}_{So_{i,j}})), 
\end{equation}
with $\mathcal{C}(\cdot)$ representing the operation of retrieving the corresponding pixel colour from the camera image. 

Finally, the dense map $\mathcal{M}$ is represented as a mesh using the Truncated Signed Distance Function (TSDF)  \cite{vizzo2022vdbfusion}. 
\begin{figure}
  \centering
  \includegraphics[width=0.48\textwidth]{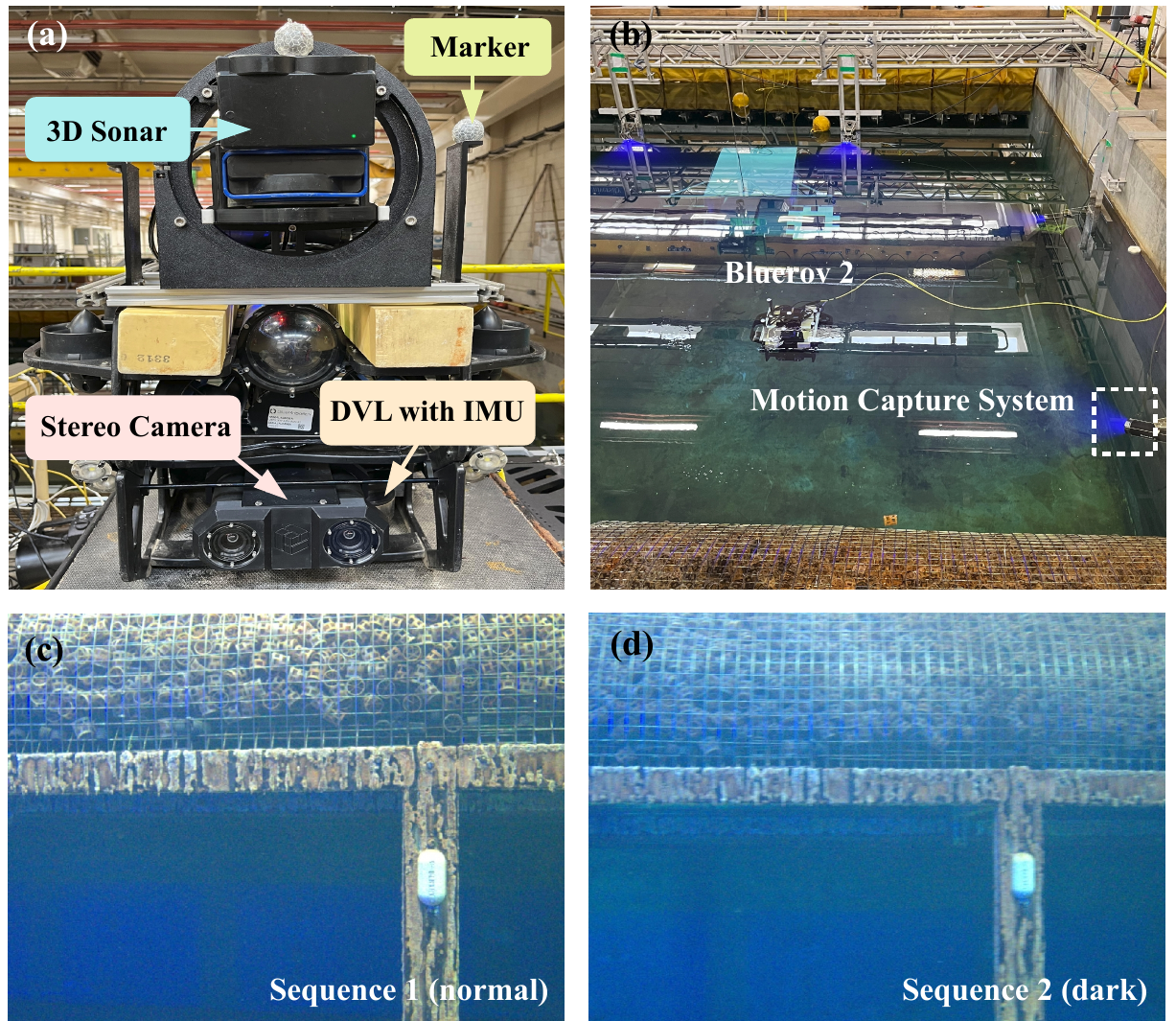}
  \caption{(a) The robot platform. (b) The laboratory tank environment with the motion capture system setup. (c) Visualisation of the tank sequence 1. (d) Visualisation of the tank sequence 2.}
  \label{fig:TankEnvironment}
\end{figure}

\begin{figure}[h]
  \centering
  \includegraphics[width=0.48\textwidth]{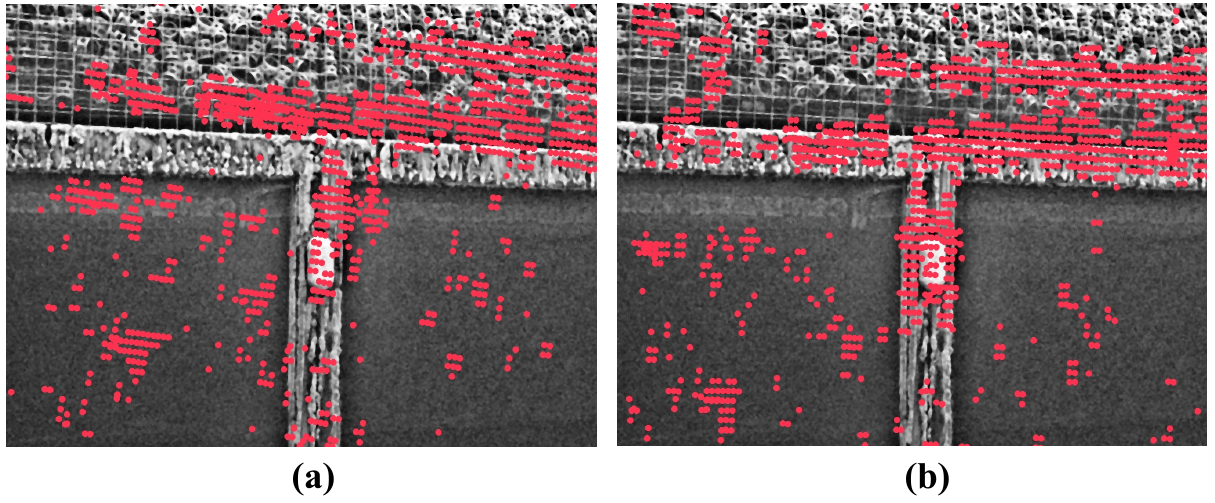}
  \caption{Back-projection of the 3D sonar point cloud into the corresponding camera image using the coarse calibration result (a) and the refined calibration result (b).}
  \label{fig:CalibrationResult}
\end{figure}

\section{EXPERIMENTAL RESULTS}\label{sec:experiment}
To validate the feasibility of the proposed VISO, we conducted extensive experimental evaluations in both a laboratory tank and an open lake. 
In all experiments, the state-of-the-art (SOTA) underwater SLAM algorithm SVIn2 \cite{rahman2022svin2} and the visual–inertial odometry algorithm VINS-Fusion \cite{qin2017vins} are used as baselines for comparison. Since a profiling sonar is prohibitively expensive and not available in our setup, only the camera and IMU are used, hence denoted as SVIn2 (VI).  Apart from the SOTA visual SLAM algorithms, we also compare against the Dead Reckoning approach \cite{moore2015generalized}, which fuses measurements from an Attitude and Heading Reference System (AHRS), a depthometer, a compass, and a DVL using an Extended Kalman Filter (EKF).

\subsection{Lab Tank Experiments}
We first conducted experiments in a laboratory tank measuring $12\times12m$ with a depth of $2.85m$. A Qualisys underwater motion capture system, equipped with four cameras mounted around the tank, was used to provide ground-truth, as shown in Fig.~\ref{fig:TankEnvironment}(b). The robot platform is a Bluerov2 with an eight-thruster configuration. It is equipped with an AHRS, a stereo camera built by Frontier Robotics, a WaterLinked Sonar 3D-15 imaging sonar, a Nortek Nucleus 1000 DVL that contains an IMU, a depth sensor, and an altitude sensor, as shown in Fig.~\ref{fig:TankEnvironment}(a). In all experiments, we use the IMU located in the Nortek Nucleus 1000 DVL instead of the onboard IMU of the Bluerov2.

We collected two datasets in the laboratory tank. The first was recorded under normal environmental lighting, while the second was with all external lights turned off, as shown in Fig.~\ref{fig:TankEnvironment}(c) and (d).

\begin{figure*}[t]
  \centering
  \includegraphics[width=0.98\textwidth]{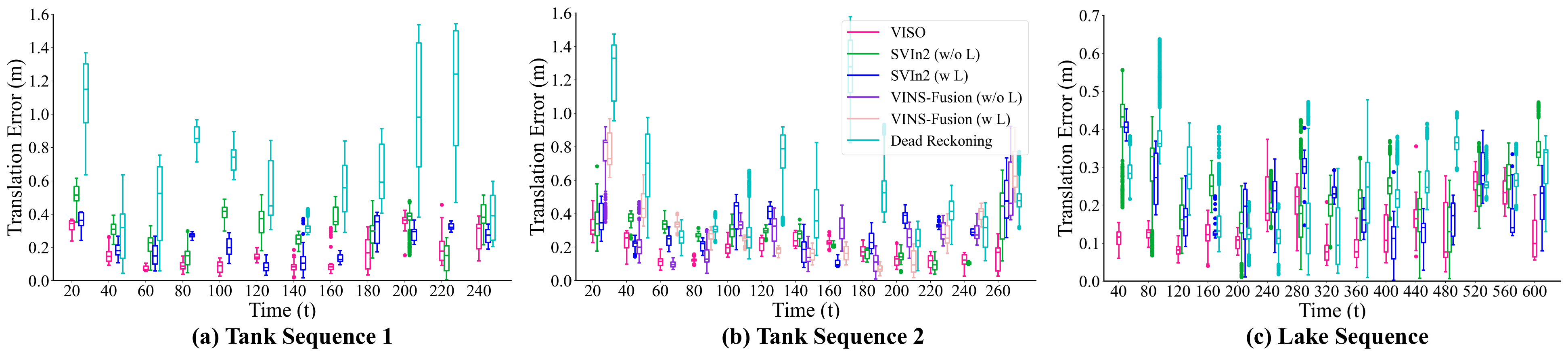}
  \caption{Translation error statistics results, where each box plot represents the translation errors over 20-second intervals in the tank sequences and 40-second intervals in the lake sequence for trajectories generated by different SLAM algorithms.}
  \label{fig:translationErrorBox}
\end{figure*}
\begin{figure*}[t]
  \centering
  \includegraphics[width=0.98\textwidth]{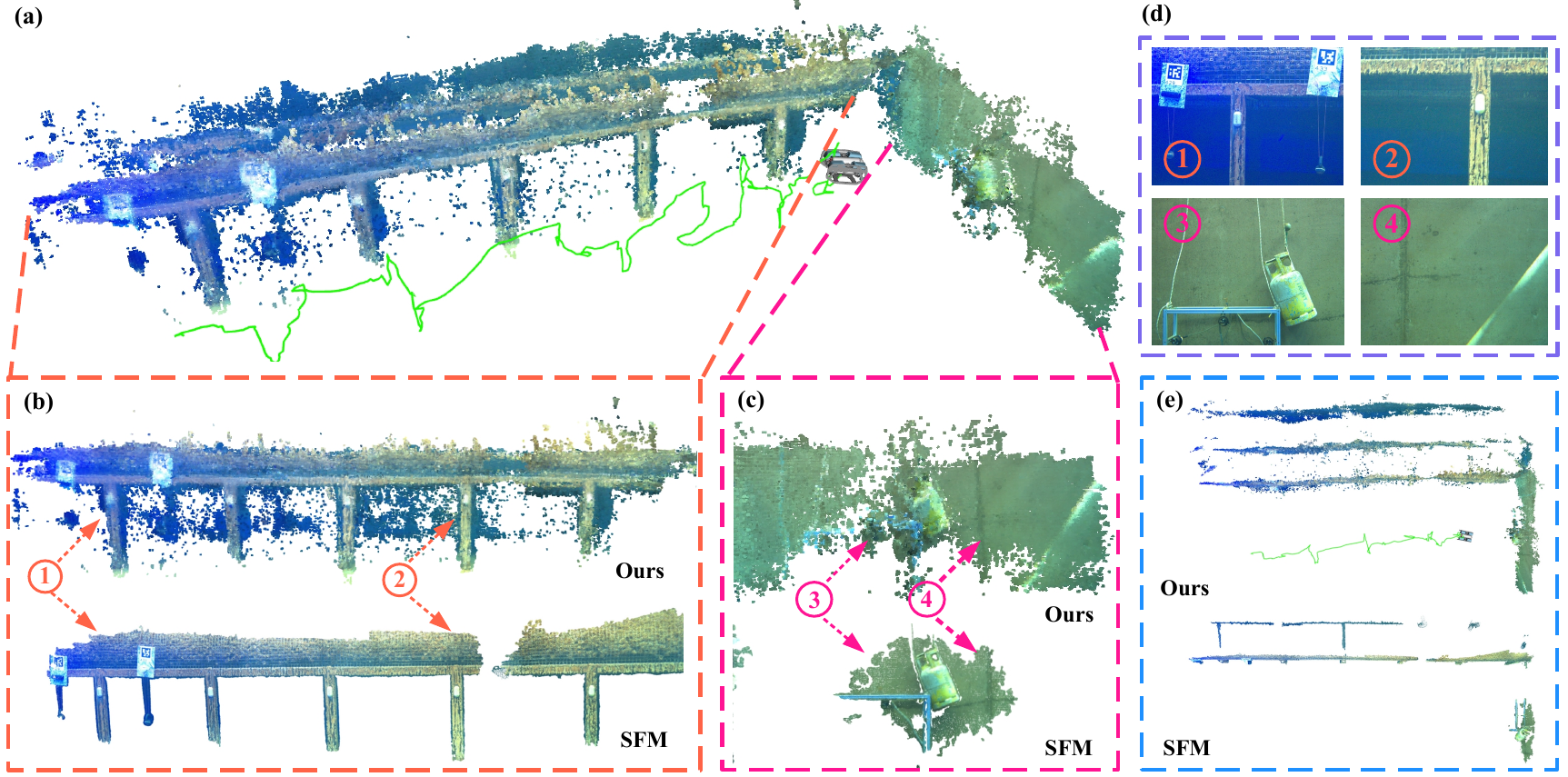}
  \caption{(a) Dense mapping result of our proposed method in the tank. (b) and (c) Front and side views of our map and SFM map, respectively. (d) Raw camera images corresponding to the regions shown in (b) and (c). (e) Top-view comparison of our method and SFM mapping results.}
  \label{fig:TankMapping}
\end{figure*}

\begin{table*}
    \footnotesize
    \centering
    \caption{Translation RMSE (m) and rotation RMSE ($^{\circ}$) across all sequences}
    \label{tab:6DOF_RMSE}
    \begin{tabular}{
        l
        |p{2.1cm}p{2cm}
        !{\vrule}
        p{2.1cm}p{2cm}
        !{\vrule}
        p{2.1cm}p{2cm}
    }
    \hline
    & \multicolumn{2}{c!{\vrule}}{\textbf{Tank Sequence 1 (normal)}} 
    & \multicolumn{2}{c!{\vrule}}{\textbf{Tank Sequence 2 (dark)}} 
    & \multicolumn{2}{c}{\textbf{Lake Sequence}} \\
    \diagbox[dir=NW, width=2.9cm]{\textbf{Method}}{\textbf{Sequences}} & Translation RMSE & Rotation RMSE  & Translation RMSE  & Rotation RMSE  & Translation RMSE & Rotation RMSE\\
    \hline
    VISO       
        & \textbf{0.201} & \textbf{5.946} & \textbf{0.213}
        & \textbf{6.140} & \textbf{0.175} & \textbf{1.554} \\
    \hline
    SVIn2 (VI) (w/o L)           
        & 0.340 & 9.424 & 0.280
        & 13.887 & 0.253 & 1.958\\
    \hline
    SVIn2 (VI) (w L)          
        & 0.249 & 20.298 & 0.315
        & 10.378 & 0.191 & 3.152 \\
    \hline
    VINS-Fusion (w/o L)   
        & N/A & N/A & 0.351
        & 11.537 & N/A & N/A \\
    \hline
    VINS-Fusion (w L)   
        & N/A & N/A & 0.382
        &9.819  & N/A & N/A
        \\
    \hline
    Dead Reckoning    
        & 0.773 & 24.99 & 0.659
        & 38.397 & 0.248 & 2.889\\
    \hline
    \end{tabular}\par
    \smallskip
    The 'N/A' indicates SLAM failed in this sequence. 'w L' and 'w/o L' indicate with and without loop closure, respectively. 
\end{table*}

\subsubsection{Online Calibration Evaluation}
First, we performed an online calibration to estimate the extrinsic parameters between the 3D sonar and the camera. Our calibration algorithm follows a coarse-to-fine process to estimate the transformation between the two sensors. Since the stereo camera and 3D sonar were mounted arbitrarily while ensuring FOV overlap, there is no ground-truth transformation available. Therefore, we present qualitative results by visualising the back-projected sonar points on the corresponding camera images to demonstrate calibration accuracy. The coarse calibration result is shown in Fig.~\ref{fig:CalibrationResult}(a), where the back-projected sonar points (red) are roughly aligned with the camera image but still exhibit noticeable misalignment. After refinement, the sonar points align well with the camera image, as shown in Fig.~\ref{fig:CalibrationResult}(b).


\begin{figure}
  \centering
  \includegraphics[width=0.48\textwidth]{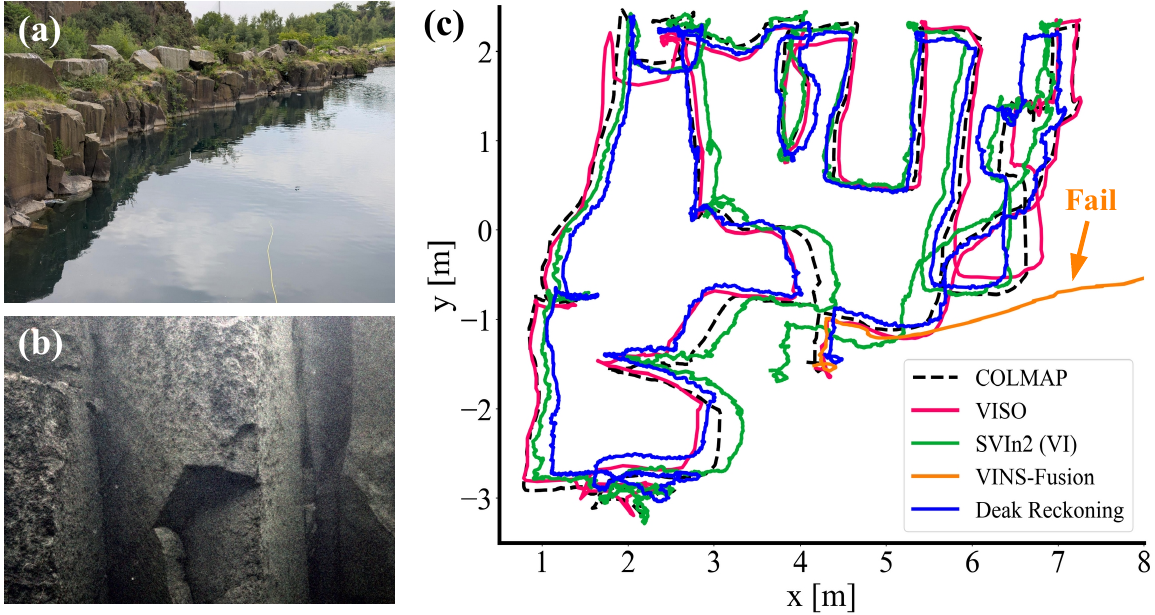}
  \caption{(a) The environment of the lake experiment. (b) The camera view in this experiment. (c) Trajectory comparison of different SLAM algorithms.}
  \label{fig:LakeTrajectory}
\end{figure}

\begin{table}
    \footnotesize
    \centering
    \caption{RMSE (m) of the localisation in ablation sequences}
    \label{tab:ablationRMSE}
    \begin{tabular}{c|p{2.6cm}|p{2.6cm}}
    \hline
    \diagbox[dir=NW, width=2.5cm]{\textbf{Method}}{\textbf{Sequences}} & Tank Sequence & Lake Sequence\\
    \hline
    VISO (w V)       &\textbf{0.170}  & \textbf{0.048}  \\
    \hline
    VISO (w/o V)        &\textcolor{gray}{\textbf{0.190}}  & \textcolor{gray}{\textbf{0.064}}  \\
    \hline
    SVIn2 (VI)       &0.192  & 0.067   \\
    \hline
    VINS-Fusion       &0.248  & 0.066   \\
    \hline
    Dead Reckoning   &0.365 &0.122 \\
    \hline
    \end{tabular}\par
    \smallskip
    The 'w V' and 'w/o V' denote with and without the camera, respectively.   
\end{table}

\subsubsection{Underwater Localization Evaluation}
We evaluate the localisation accuracy in the two tank sequences. For a more comprehensive comparison, VISO is compared alongside SVIn2 and VINS-Fusion under both loop closure and non–loop closure settings.
The qualitative results are shown in Fig.~\ref{fig:translationErrorBox}(a) and (b), where the absolute translation errors over each 20 seconds are plotted.
The figure shows that VISO has lower translation errors overall across both tank sequences, and the smaller box height, which represents the standard deviation of error, highlights the robustness of our approach. Additionally, we provide a statistical evaluation of the translation and rotation performance of all SLAM algorithms across the two tank sequences.
The quantitative results are summarised in Table~\ref{tab:6DOF_RMSE}, which shows that VISO consistently outperforms competing algorithms in both translation and rotation accuracy, further demonstrating its superiority in localisation.

We also evaluated the robustness of VISO with 3D sonar integration under visual degradation. Specifically, we compared the performance of VISO with and without stereo camera data against baseline algorithms. For this experiment, we used the first 120 seconds of data in tank sequence 2, as the later portion included a brief period of 3D sonar degradation caused by objects moving out of range. The quantitative localisation errors are summarised in Table~\ref{tab:ablationRMSE}, which illustrates that VISO achieves the best overall performance using visual data. Notably, it remains robust and highly accurate even with the camera disabled, outperforming other SOTA algorithms and demonstrating strong reliability in visually challenging environments.


\subsubsection{Dense Mapping Experiment}
In this work, we propose a real-time novel dense mapping approach for underwater 3D scene reconstruction using 3D sonar data and photometric rendering. To evaluate its performance, we conducted a mapping experiment in the laboratory tank and compared our approach with COLMAP \cite{schonberger2016structure}, a widely used structure-from-motion (SFM) method for offline dense mapping. As shown in Fig.~\ref{fig:TankMapping}, our method, which exploits rendered 3D sonar point clouds for dense mapping, achieves performance comparable to the SOTA offline algorithm. However, while SFM requires approximately 20 minutes on a server equipped with 8 TITAN X GPUs and a 32-core Intel(R) Xeon(R) CPU, our method generates the map in real time.

In addition, our method is more efficient for dense mapping as it does not require revisiting the same location. As shown in Fig.~\ref{fig:TankMapping}(c), the SFM map loses the peripheral information compared to ours due to the lack of revisits in that area. Moreover, the 3D sonar provides absolute range measurements using acoustics, making it more robust for depth estimation in visually challenging environments, as it does not rely on triangulation or stereo matching. Furthermore, acoustic signals can penetrate structures and capture environmental details that cameras cannot perceive, as illustrated in Fig.~\ref{fig:TankMapping}(e), allowing our map to provide richer environmental information that is crucial for autonomous vehicle motion planning.

\subsection{Open Lake Experiments}
We further conducted an experiment in an open lake to evaluate the localisation and mapping performance of VISO in a complex environment, as shown in Fig.~\ref{fig:LakeTrajectory}(a). Since no ground-truth is available in the lake, we use the trajectory generated by COLMAP as a reference. All SLAM trajectories are compared with the COLMAP trajectory for both qualitative and quantitative localisation evaluation. In this experiment, VINS-Fusion failed to operate over the entire sequence due to illumination variations, as illustrated in Fig.~\ref{fig:LakeTrajectory}(b). The trajectory comparison is shown in Fig.~\ref{fig:LakeTrajectory}(c). For a more detailed evaluation, the absolute translation errors over each 40 seconds are plotted in Fig.~\ref{fig:translationErrorBox}(c), while the quantitative results are provided in Table \ref{tab:6DOF_RMSE}. Both qualitative and quantitative results indicate that VISO outperforms the other algorithms in localisation accuracy.

Additionally, we achieve high-reality underwater dense mapping using 3D point clouds and photometric rendering, as shown in Fig.~\ref{fig:LakeMapping}(a). The dense map closely matches the actual scene, as highlighted by the colored dotted boxes in Fig.~\ref{fig:LakeMapping}(a) and the corresponding camera images in Fig.~\ref{fig:LakeMapping}(b). 

\begin{figure}
  \centering
  \includegraphics[width=0.48\textwidth]{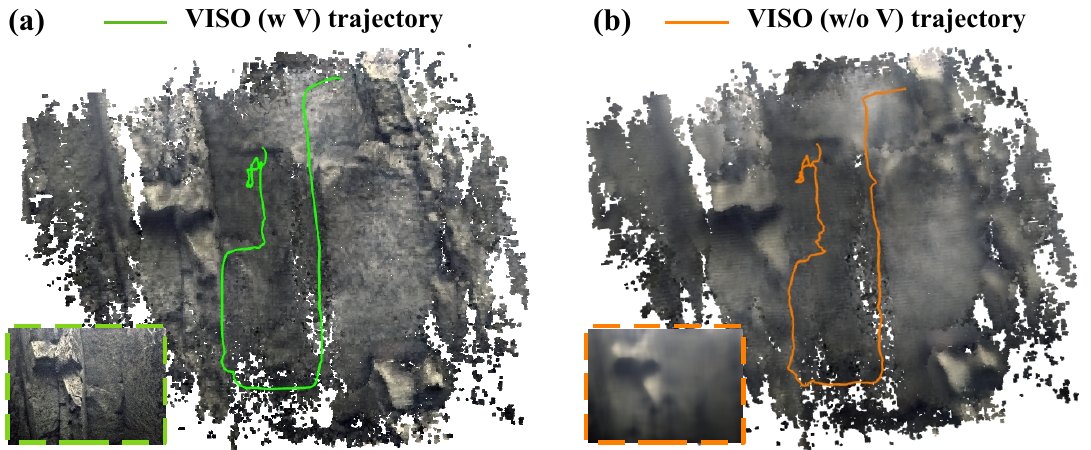}
  \caption{The mapping comparison of VISO with (w V) and without (w/o V) camera residual. The dotted boxes are camera view.}
  \label{fig:LakeAblationMapping}
\end{figure}

To assess the robustness and accuracy of VISO in visually challenging environments, we conducted an ablation experiment. A 90-second segment from the lake sequence was selected to ensure that VINS-Fusion could operate. In this experiment, we disabled the camera residual of VISO and blurred the camera images to simulate high turbidity, as indicated by the dotted box in Fig.~\ref{fig:LakeAblationMapping}(b). The mapping results of VISO with and without camera residual are shown in Fig.~\ref{fig:LakeAblationMapping}(a) and (b), respectively, demonstrating its dense mapping capability in both clear and turbid environments. We also compare the localisation ability of VISO with other SOTA algorithms, with the results summarised in Table~\ref{tab:ablationRMSE}. Since the visual conditions in this dataset are favourable, all algorithms achieved high localisation accuracy. Nevertheless, VISO with camera data outperformed the baselines, while VISO without camera achieved accuracy comparable to visual odometry, highlighting the potential of our proposed SLAM system in visually impaired environments.

\section{Conclusion}\label{sec:conclusion}
This work presents an underwater SLAM system, VISO, which fuses a stereo camera, an IMU, and a 3D sonar to achieve robust localisation and highly realistic dense 3D reconstruction in underwater environments. Extensive experiments in both a laboratory tank and an open lake demonstrate that integrating the 3D sonar not only enhances robustness and accuracy in visually challenging conditions but also enables real-time dense mapping using 3D sonar point clouds. In particular, fusing 3D sonar data with camera information allows the 3D sonar point cloud map to be rendered with photometric information, which is valuable for underwater applications such as inspection and navigation.


\bibliographystyle{IEEEtran}
\bibliography{reference}

\end{document}